\documentclass[10pt,twocolumn,letterpaper]{article}

\usepackage[pagenumbers]{cvpr} 

\usepackage{colortbl}

\definecolor{cvprblue}{rgb}{0.21,0.49,0.74}
\usepackage[pagebackref,breaklinks,colorlinks,allcolors=cvprblue]{hyperref}

\def\confName{CVPR}
\def\confYear{2027}

\title{ALERT: Extending Humanoid Motion Trackers with a General and Lightweight Human Avoidance Pipeline}

\author{
\begin{tabular*}{0.70\textwidth}[b]{@{\extracolsep{\fill}}ccc@{}}
Zhaoyang Li$^{1,2}$\thanks{Equal contribution.} &
Chiyu Zhang$^{1,2*}$ &
Chengxin Xu$^{3*}$
\end{tabular*}\\[6pt]
\begin{tabular*}{0.94\textwidth}[b]{@{\extracolsep{\fill}}cccc@{}}
Peng Yang$^{1,2}$ &
Zhonghao Guo$^{1,2}$ &
Wanting Li$^{1,2}$\thanks{Corresponding authors: Wanting Li (\texttt{wanting.li@ia.ac.cn}) and Ziyu Chen (\texttt{ziyu.chen@ia.ac.cn}).} &
Ziyu Chen$^{1,2\dagger}$
\end{tabular*}\\[10pt]
$^{1}$Institute of Automation, Chinese Academy of Sciences\\[2pt]
$^{2}$University of Chinese Academy of Sciences\hspace{1.5em}%
$^{3}$Beijing Institute of Technology
}

\begin{document}
\maketitle
\begin{abstract}
Whole-body motion tracking has enabled humanoid dancing, athletic motions, and teleoperation, providing a foundation for capability demonstrations and manipulation data collection. However, existing tracking policies lack awareness of nearby humans and proactive avoidance capabilities, while motion substitution and local constraint-based corrections may compromise reference motion integrity and tracking continuity. We propose ALERT, a general lightweight pipeline for extending motion tracking to shared human–robot environments. Through lightweight adaptation, it integrates human perception and proactive avoidance into pretrained tracking policies, balancing avoidance demands with reference motion tracking while preserving existing motion capabilities. Simulation and real-world experiments across tracking backbones demonstrate reliable collision-free execution through localized human avoidance, preserving whole-body coordination and enabling rapid recovery of reference tracking, while retaining the ability of pretrained policies to track unseen motions.
\end{abstract}    
\section{Introduction}
\label{sec:intro}
\begin{figure}[t]
  \centering
  \includegraphics[width=\linewidth]{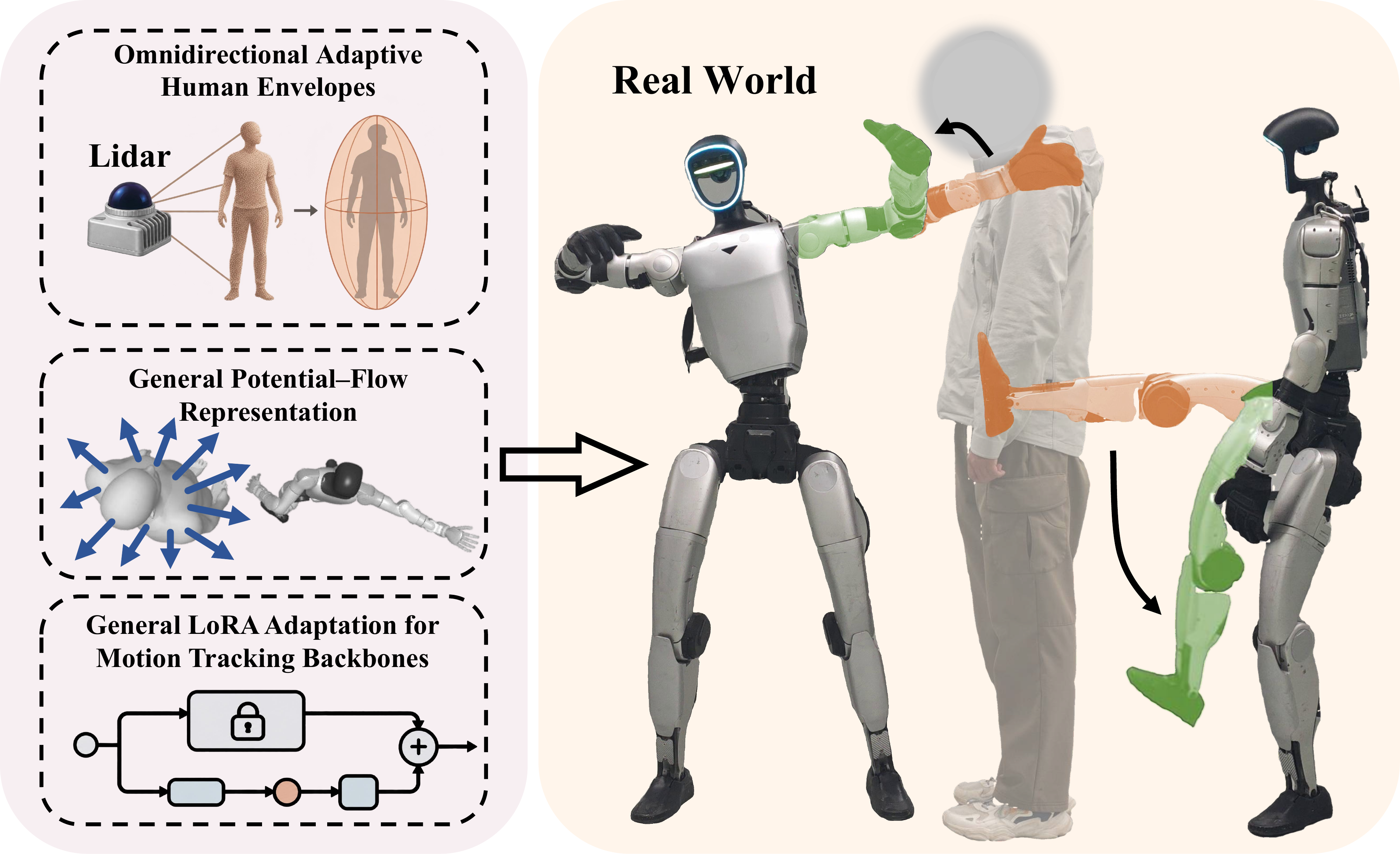}
  \caption{\textbf{ALERT extends motion trackers with human avoidance.}
  LiDAR-based human envelopes, potential flows, and LoRA (left) enable local avoidance on hardware (right).
  Orange and green show reference and adjusted poses, respectively.}
  \label{fig:teaser}
\end{figure}

Recent advances in reinforcement learning for whole-body motion tracking have substantially improved humanoid motion capabilities,
enabling robots to reproduce complex human motions such as dance and athletic movements on real hardware
and perform real-time whole-body teleoperation~\cite{luo2026sonic,ze2026twist2,zeng2026behavior}.
These capabilities support robot performances, manipulation data collection, and human skill transfer.
However, existing motion-tracking policies generate control commands primarily from proprioceptive states and reference motions,
without perceiving or actively avoiding nearby people.
Arm swings and kicks may injure nearby spectators,
while on-site personnel may enter the robot limb workspace during teleoperation.
Deploying these capabilities in shared human--robot settings therefore requires
human-aware motion tracking that accommodates nearby people throughout whole-body motion execution.

Mainstream pretrained trackers provide strong tracking capabilities across diverse motions
through learned motion representations and coordinated whole-body control,
which should be fully utilized when learning human avoidance.
Meanwhile, large-scale joint retraining for motion tracking and human avoidance requires substantial data and computational resources.
This motivates a lightweight adaptation pipeline that extends pretrained trackers with human avoidance
rather than relearning whole-body control.

The challenge is to add human avoidance while preserving pretrained motion-tracking capabilities.
Corrections should remain small and brief, concentrate on body parts at risk,
and allow reference tracking to resume as the risk subsides.
Each correction changes subsequent tracking states,
coupling avoidance with whole-body coordination and support contacts.
CSF~\cite{yang2026csf} uses visual scenes and language rules to replace unsafe references with safe continuations,
sacrificing original motion content and continuity.
ConstrainedMimic~\cite{morton2026constrained} uses control barrier functions
and single-step constrained optimization to modify reference motions and low-level control outputs.
However, corrections satisfying current geometric constraints can disrupt balance and subsequent tracking
when avoidance requires longer-horizon coordination or contact transitions.
These limitations call for learning avoidance directly within the control policy to coordinate avoidance and tracking over longer horizons.

To this end, we propose \textbf{ALERT} (\textbf{A}daptive \textbf{L}ow-Rank \textbf{E}xtension Pipeline for \textbf{R}eference \textbf{T}rackers),
a general lightweight training pipeline that extends pretrained motion trackers with human avoidance
while preserving their motion-tracking capabilities.
We freeze the pretrained backbone and introduce potential-conditioned low-rank branches into its action network
to learn local avoidance corrections using existing whole-body coordination.
Each tracker retains its native motion-conditioning mechanism,
while a shared potential-field representation and velocity-flow training objective
make the pipeline applicable across mainstream tracking frameworks.
We use omnidirectional LiDAR to detect and track nearby humans
and estimate their occupied space with convex envelopes that adapt to body shape and pose.
Superposing repulsive potentials from these envelopes provides a fixed-size representation
of the surrounding avoidance demands across varying human numbers and arrangements.
To coordinate avoidance with reference tracking,
we combine potential-guided avoidance velocities with the reference motion flow
and train the adaptation branches to track the resulting target flow.
This guidance preserves ongoing motion trends, limits interference with unaffected limbs,
and recovers the reference flow as clearance increases.
As illustrated in Fig.~\ref{fig:teaser}, ALERT adjusts an arm or leg to avoid nearby people
while preserving the overall reference pose on real hardware.

Using only $500$ motion sequences that intrude into the surrounding space and a single NVIDIA RTX 4090 GPU,
we extend multiple mainstream tracking frameworks with human avoidance.
The adapted policies largely retain their original tracking accuracy without nearby humans
and concentrate avoidance deviations on limbs at risk when people approach.
On real hardware, the robot avoids nearby people during dance and kicking, then rapidly resumes reference tracking.
During teleoperation, affected limbs adjust autonomously while the rest of the body follows the operator.
Compared with CSF and ConstrainedMimic, ALERT completes more motions without collisions or falls
and achieves lower tracking errors throughout avoidance and recovery.

Our contributions are fourfold:
\begin{itemize}
  \item We develop a general lightweight training pipeline that adds human avoidance to pretrained trackers
  while preserving their motion-tracking capabilities.
  \item We combine LiDAR-based adaptive human envelopes with potential superposition
  to represent varying human numbers and arrangements using fixed-size inputs.
  \item We formulate potential-guided target velocity flows and a tracking reward
  that guide local avoidance while maintaining the continuity of reference motion.
  \item We validate sim-to-real transfer through human avoidance and motion preservation
  during real-world dance, kicking, and whole-body teleoperation.
\end{itemize}

\section{Related Work}

\subsection{Learning-Based Whole-Body Control}

Learning-based whole-body control enables humanoids to reproduce diverse human motions.
ExBody and ExBody2~\cite{cheng2024expressive,ji2026exbody2} learn expressive control from human motion data,
while WoCoCo~\cite{zhang2024wococo} learns control with sequential contacts.
DreamControl~\cite{kalaria2026dreamcontrol} uses guided diffusion for scene interaction.
Whole-body controllers and teleoperation systems~\cite{ze2026twist2,he2025hover,he2024omnih2o,ben2025homie}
connect human motion inputs to learned policies for real-time imitation and manipulation data collection.
SONIC~\cite{luo2026sonic} and BFM-Zero~\cite{li2026bfm} scale motion tracking and reusable behavior representations
across diverse skills.
However, these controllers primarily follow reference motions and lack proactive avoidance of nearby people during execution.
ALERT extends these pretrained trackers through a lightweight training pipeline, enabling human avoidance at low adaptation cost.

\subsection{Safe Collision Avoidance for Humanoid Robots}

Collision avoidance for humanoids spans local planning and whole-body control.
Prediction-aware potential fields~\cite{guo2026prediction} and vision-based motion planning~\cite{shang2025adaptive,krishna2026vibe}
anticipate moving obstacles in human--robot collaboration.
Control barrier functions impose whole-body safety constraints~\cite{khazoom2022humanoid,chen2025control},
while recent methods explore constrained tracking, collision-aware policy learning,
and generative safety filtering~\cite{morton2026constrained,gadde2026collision,yi2026swap,yang2026csf}.
Potential fields also guide humanoid traversal through cluttered scenes~\cite{xue2026collision}.
However, geometric clearance alone does not specify how to adjust body motion while continuing reference tracking.
ALERT superposes potentials from adaptive human envelopes to form a general representation
across varying human counts, shapes, and arrangements.
The resulting velocity-flow tracking reward provides dense, directional guidance before contact,
coordinating local avoidance with the ongoing reference motion.

\subsection{Adaptation of Pretrained Control Policies}

Policy adaptation extends learned skills to new tasks and dynamics.
Residual reinforcement learning~\cite{johannink2019residual} adds corrections to base controllers,
while Robot Trains Robot~\cite{hu2025robot} adapts dynamics representations on hardware.
Humanoid applications include loco-manipulation~\cite{zhao2025resmimic},
fast task adaptation~\cite{wang2026general}, long-horizon tracking~\cite{sun2026robotdancing},
and sim-to-real alignment~\cite{he2025asap}.
AdaMimic~\cite{huang2026towards} adapts tracking speed and low-level actions,
while ADP~\cite{chen2025adp} uses low-rank updates to adapt diffusion policies.
Rapid motor adaptation~\cite{kumar2021rma}, simulation-guided fine-tuning~\cite{yin2025rapidly},
and meta-reinforcement learning~\cite{nagabandi2018learning} address changing deployment conditions.
ALERT introduces potential-conditioned low-rank branches into the frozen action network,
learning local avoidance corrections while preserving pretrained tracking skills and whole-body coordination.

\begin{figure*}[t]
  \centering
  \includegraphics[width=\textwidth]{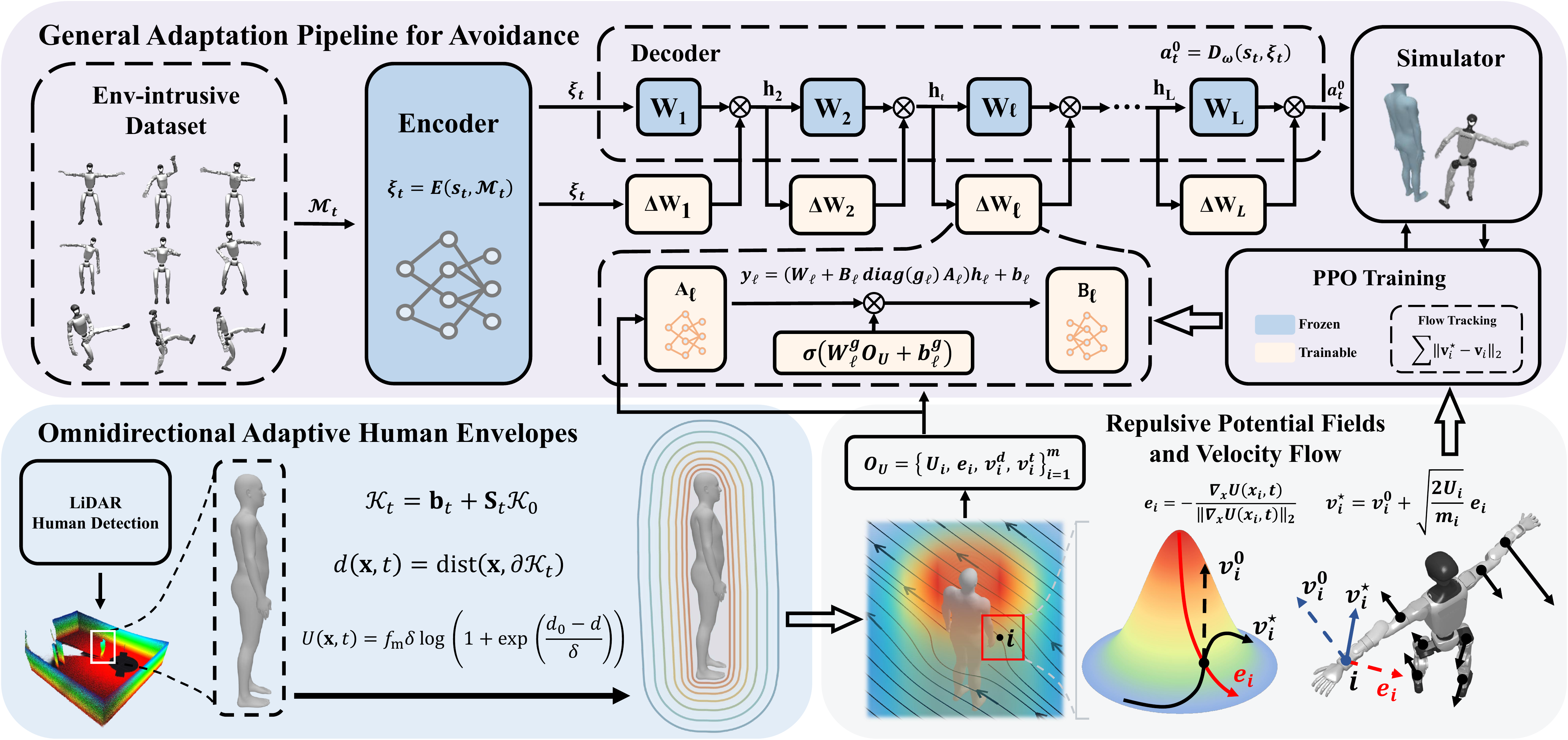}
  \caption{Overview of ALERT: Extending Humanoid Motion Trackers with a General and Lightweight Human Avoidance Pipeline.}
  \label{fig:method_overview}
\end{figure*}

\section{Method}
\label{sec:method}

As shown in Figure~\ref{fig:method_overview}, ALERT extends a pretrained whole-body motion-tracking policy
with conditional low-rank adaptation for human avoidance (Sec.~\ref{sec:df_common_framework}).
We estimate adaptive human envelopes from real-time point clouds (Sec.~\ref{sec:human_obstacle_field})
and use them to construct repulsive potential fields.
The resulting avoidance flow is added to the reference motion flow to form a target velocity flow (Sec.~\ref{sec:df_repulsive_power}).
Guided by this target flow, the policy learns local motion adjustments conditioned on potential-field features (Sec.~\ref{sec:df_conditioned_lora}).

\subsection{General Adaptation Pipeline for Avoidance}
\label{sec:df_common_framework}

Existing motion trackers differ in reference representations and network architectures,
but share two control stages: motion condition construction and joint action generation.
Let $\mathbf s_t$ denote the proprioceptive state,
$\mathcal M_t$ the reference motion or target information,
and $E$ the condition construction operator of the base policy.
The pretrained policy is then written as:
\begin{equation}
  \boldsymbol\xi_t=E(\mathbf s_t,\mathcal M_t),
  \qquad \mathbf a_t^0=D_\omega(\mathbf s_t,\boldsymbol\xi_t).
  \label{eq:df_common_tracker}
\end{equation}
Here, $\boldsymbol\xi_t$ specifies the motion condition,
and $D_\omega$ combines it with the current proprioceptive state to produce the joint-action mean $\mathbf a_t^0$.
This formulation encompasses multiple backbones.
In SONIC~\cite{luo2026sonic}, $\boldsymbol\xi_t$ is an encoded and quantized motion representation,
and $D_\omega$ is the control decoder.
In TWIST2~\cite{ze2026twist2}, $\boldsymbol\xi_t$ is a tracking condition constructed from whole-body reference motions,
and $D_\omega$ is the whole-body motion-tracking policy.
In BFM-Zero~\cite{li2026bfm}, $\boldsymbol\xi_t$ is a goal-conditioned behavior latent,
and $D_\omega$ is the action decoder.

Under this abstraction, local human avoidance is formulated as potential-field-conditioned adaptation of the action mapping $D_\omega$.
The reference condition $\boldsymbol\xi_t$ specifies the motion content,
the proprioceptive state $\mathbf s_t$ describes the current execution state,
and the potential-field representation $O_U$ provides local potential energy, descent directions, and flow information for each body part.
The adapted policy $D_{\omega,\theta}(\mathbf s_t,\boldsymbol\xi_t;O_U)$ combines these inputs
to adjust body parts near humans while continuing the reference motion.
Each backbone retains its own condition construction and learns avoidance from a common potential-field input and target velocity flow.

Specifically, we adopt conditionally gated low-rank adaptation (LoRA)~\cite{hu2021lora,zindari2026learning}
to introduce potential-field-conditioned weight increments into the linear layers of the action network:
\begin{equation}
  \mathbf y_\ell
    =\big(W_\ell+B_\ell\operatorname{diag}(\mathbf g_\ell)A_\ell\big)
      \mathbf h_\ell+\mathbf b_\ell.
  \label{eq:df_gated_lora}
\end{equation}
Here, $\mathbf h_\ell$ is the hidden-layer input, while $W_\ell$ and $\mathbf b_\ell$ remain frozen.
The matrices $A_\ell$ and $B_\ell$ form an increment of rank at most $r$,
modulated by the gating vector $\mathbf g_\ell=\sigma(G_\ell O_U+\mathbf b_\ell^g)$,
where $\sigma$ is the sigmoid function.
The adaptation parameters $\theta$ comprise the low-rank projections and gating parameters, and $B_\ell$ is initialized to zero.

The pretrained policy provides whole-body coordination and motion-tracking capabilities,
while fine-tuning focuses on learning local action corrections near humans.
In our experiments, fine-tuning on a relatively small set of motion sequences enables avoidance across all three backbones
while preserving their original tracking performance.

\subsection{Omnidirectional Adaptive Human Envelopes}
\label{sec:human_obstacle_field}

Avoiding nearby humans requires continuous and reliable perception of human positions and occupied space.
However, LiDAR measurement noise and human self-occlusion introduce local jitter and missing observations,
making it difficult to reliably delineate a safe envelope for limb motion from a single frame.
To address this issue, we combine temporal human tracking with parametric convex-body modeling
to estimate the human center trajectory and outer envelope, respectively.

For human tracking, we first detect humans and extract a reference point cloud using Person-MinkUNet~\cite{jia2021person}.
A particle filter~\cite{elfring2021particle} then registers this reference cloud to the latest observations around the target to estimate its current position,
while an $\alpha$--$\beta$ position--velocity filter~\cite{khan2024optimizing} produces a smooth, continuous center trajectory.
Let $\mathcal K_0$ denote a canonical convex template whose base lies on $z=0$ and whose horizontal center is at the origin,
and let $\mathbf b_t$ denote the three-dimensional translation determined by the tracked center and ground height.
The human envelope $\mathcal K_t$ is:
\begin{equation}
  \mathcal K_t=\mathbf b_t+\mathbf S_t\mathcal K_0,
  \label{eq:df_scaled_convex_envelope}
\end{equation}
where $\mathbf S_t=\operatorname{diag}(s_{x,t},s_{y,t},s_{z,t})$ is a positive diagonal scaling matrix.
By updating the three axis scales according to the point cloud distribution,
the template adapts to changes in occupied space across body shapes and poses.
We update $\mathbf S_t$ online so that the convex envelope $\mathcal K_t$ satisfies the coverage constraint
$|\mathcal X_t\cap\mathcal K_t|/M_t\ge 1-\varepsilon$,
where $\mathcal X_t$ is the target point cloud containing $M_t$ observed points.
The outlier tolerance $\varepsilon$ allows a small fraction of noisy points to lie outside the envelope,
reducing their influence on its scale.

To make the repulsive potential vary smoothly with the human envelope,
we define the distance~\cite{oleynikova2017voxblox} from a query point $\mathbf x\notin\mathcal K_t$ to the envelope boundary:
\begin{equation}
  d(\mathbf x,t)=\operatorname{dist}(\mathbf x,\partial\mathcal K_t).
  \label{eq:df_envelope_distance}
\end{equation}
where $\operatorname{dist}(\cdot,\cdot)$ denotes Euclidean distance and $\partial\mathcal K_t$ is the envelope boundary.
This distance serves as the spatial parameter of the potential function defined below,
and its gradient provides an analytical expression for the potential descent direction.
The gradient is obtained from the nearest-point projection $\Pi_{\mathcal K_t}(\mathbf x)$ onto the convex-body surface:
\begin{equation}
  \nabla_{\mathbf x}d(\mathbf x,t)
    =\frac{\mathbf x-\Pi_{\mathcal K_t}(\mathbf x)}{d(\mathbf x,t)}.
  \label{eq:df_surface_normal}
\end{equation}
This formulation converts discrete, noisy human point clouds into spatial envelopes that are continuously updated with human motion.
In the next section, we use the envelope distance and its gradient to construct an artificial repulsive potential field.

\subsection{Repulsive Potential Fields and Velocity Flow}
\label{sec:df_repulsive_power}

During motion tracking, potential-field guidance must be coordinated with limb flow directions.
We construct an avoidance flow along the direction of steepest potential descent
and map local potential energy to the equivalent kinetic energy of a flow increment.
We add this increment to the reference motion flow
to obtain a target velocity flow at whole-body sample points for policy learning.

For a point $\mathbf x\notin\mathcal K_t$, let $d=d(\mathbf x,t)$ denote its distance to the human envelope.
Let $\rho(d)\ge0$ be the scalar field strength along the distance direction.
Taking the potential energy at infinity as zero, we define the spatial repulsive potential field as
$U(\mathbf x,t)=\int_d^{+\infty}\rho(s)\,\mathrm ds$~\cite{pan2021improved}.
As the query point approaches the human envelope, $d$ decreases and the integration interval expands,
so the local potential energy increases monotonically.
To keep the field strength bounded and smooth with respect to distance, we define:
\begin{equation}
  \rho(d)=f_{\max}\sigma\!\left(\frac{d_0-d}{\delta}\right),
  \label{eq:df_field_strength}
\end{equation}
where $\sigma(\cdot)$ is the standard sigmoid function and $f_{\max}>0$ bounds the field strength.
The parameters $d_0$ and $\delta>0$ control the center and width of the distance transition, respectively.
Substituting Eq.~\eqref{eq:df_field_strength} into the potential integral yields the closed-form expression:
\begin{equation}
  U(\mathbf x,t)=f_{\max}\delta
    \log\!\left(1+\exp\!\left(\frac{d_0-d}{\delta}\right)\right).
  \label{eq:df_spatial_potential}
\end{equation}
Since $\partial U/\partial d=-\rho(d)$, we have $-\nabla_{\mathbf x}U=\rho(d)\nabla_{\mathbf x}d$.
For a single human, $\rho(d)>0$ and $\|\nabla_{\mathbf x}d\|=1$, so the direction of steepest potential descent simplifies to:
\begin{equation}
  \mathbf e=-\frac{\nabla_{\mathbf x}U}{\|\nabla_{\mathbf x}U\|}
  =\frac{\mathbf x-\Pi_{\mathcal K_t}(\mathbf x)}{d(\mathbf x,t)}.
  \label{eq:df_descent_direction}
\end{equation}
For multiple humans, we sum the individual repulsive potentials as $U=\sum_j U^{(j)}$
and normalize the negative gradient of the total potential wherever the gradient is nonzero.

We select $m$ sample points at robot joints and link midpoints.
For the $i$-th sample point $\mathbf x_i$, let $U_i=U(\mathbf x_i,t)$ denote its local potential energy
and $\mathbf e_i=\mathbf e(\mathbf x_i,t)$ its unit steepest-descent direction.
We use an isotropic virtual mass matrix $\mathbf M=m_i\mathbf I_3$ with $m_i>0$
to scale the mapping from potential energy to the flow increment.
Let $\alpha_i\ge0$ be the increment magnitude along $\mathbf e_i$.
We map the local potential energy to the equivalent kinetic energy of the increment:
\begin{equation}
  \tfrac12(\alpha_i\mathbf e_i)^\top
    \mathbf M(\alpha_i\mathbf e_i)=U_i.
  \label{eq:df_potential_kinetic_energy}
\end{equation}
Using $\mathbf e_i^\top\mathbf M\mathbf e_i=m_i$, we obtain the nonnegative increment magnitude at each point.
Adding these increments along the steepest-descent directions to the nominal motion flow $\mathbf v^0\in\mathbb R^{3m}$
yields the whole-body target velocity flow:
\begin{equation}
  \mathbf v^\star=\mathbf v^0+
  \left[
    \sqrt{\frac{2U_1}{m_1}}\,\mathbf e_1^\top,
    \ldots,
    \sqrt{\frac{2U_m}{m_m}}\,\mathbf e_m^\top
  \right]^\top.
  \label{eq:df_workspace_velocity_flow}
\end{equation}
The target flow preserves the nominal motion component tangent to equipotential surfaces.
As local potential energy approaches zero, it recovers the nominal flow.
In the next section, this flow guides conditional low-rank adaptation
to learn local avoidance with whole-body coordination.

\subsection{Potential-Conditioned Policy Learning}
\label{sec:df_conditioned_lora}

We train the conditional LoRA branches using Proximal Policy Optimization (PPO)~\cite{schulman2017proximal},
with a Gaussian policy that uses the adapted network output as its mean and a fixed standard deviation.
We freeze the pretrained backbone and update only the LoRA parameters and value network.

Each adaptation module receives hidden-layer features $\mathbf h_\ell$ encoding the proprioceptive state and reference motion,
while $O_U$ modulates the gates of the low-rank branch.
Let $\mathbf v_i$ denote the current flow at sample point $i$.
Its components along $\mathbf e_i$ and tangent to the equipotential surface are
\mbox{$v_i^d=\mathbf e_i^\top\mathbf v_i$} and \mbox{$\mathbf v_i^t=\mathbf v_i-v_i^d\mathbf e_i$}, respectively.
We concatenate the potential and flow features:
\begin{equation}
  O_U=\big\{\,U_i,\,\mathbf e_i,\,v_i^d,\,\mathbf v_i^t\,\big\}_{i=1}^{m}.
  \label{eq:df_condition_field_features}
\end{equation}
Local potential energy and steepest-descent directions characterize repulsion,
while the flow components distinguish motion away from, toward, or around humans.

We use \mbox{$r_t=r_t^{\mathrm{base}}+r_t^{\mathrm{flow}}+r_t^{\mathrm{coll}}$} for each backbone,
combining the original reward with flow-tracking and collision terms.
The flow-tracking reward penalizes the flow-vector error at each point:
\begin{equation}
  r_t^{\mathrm{flow}}
    =-\sum_{i=1}^{m}w_i
      \left\|\mathbf v_{i,t+1}-\mathbf v_{i,t}^{\star}\right\|_2^2,
  \label{eq:df_flow_tracking_reward}
\end{equation}
where $w_i\ge0$ weights point $i$.
We set \mbox{$r_t^{\mathrm{coll}}=-10$} for human collisions and $0$ otherwise.

Potential-conditioned LoRA learns local corrections from target velocity flows.
The adapted policy balances human avoidance with whole-body reference tracking.

\section{Experiments}
\label{sec:experiments}

\begin{table*}[t]
  \centering
  \caption{Adaptation performance across three motion-tracking backbones.}
  \label{tab:cross_backbone}
  \small
  \setlength{\tabcolsep}{3pt}
  \renewcommand{\arraystretch}{1.10}
  \newcommand{\metricstat}[2]{\ensuremath{#1_{\pm #2}}}
  \begin{tabular*}{\textwidth}{@{\extracolsep{\fill}}lccccccc@{}}
    \toprule
    Method
      & $E_{\mathrm{g\text{-}mpbpe}}$
      & $E_{\mathrm{mpbpe}}^{a}$
      & $E_{\mathrm{mpbpe}}^{o}$
      & $E_{\mathrm{mpboe}}$
      & $E_{\mathrm{mpjpe}}$
      & $E_{\mathrm{mpjve}}$
      & $\mathcal{M}$ \\
    \midrule
    \rowcolor{black!7}
    \multicolumn{8}{l}{\textbf{SONIC}} \\
    Original
      & \metricstat{52.32}{4.92} & \metricstat{37.29}{0.42} & \metricstat{35.87}{1.79} & \metricstat{182.73}{3.31}
      & \metricstat{113.54}{2.02} & \metricstat{6.47}{0.11} & \metricstat{0.00}{0.00} \\
    Ours (w/o H)
      & \metricstat{54.70}{4.80} & \metricstat{40.16}{0.97} & \metricstat{37.36}{1.97} & \metricstat{191.64}{5.66}
      & \metricstat{123.33}{2.69} & \metricstat{6.59}{0.11} & / \\
    Ours (w/ H)
      & \metricstat{62.85}{6.33}
      & \metricstat{84.03}{19.30}
      & \metricstat{37.75}{7.51}
      & \metricstat{204.30}{3.61}
      & \metricstat{136.21}{2.43}
      & \metricstat{6.69}{0.12}
      & \metricstat{1.21}{0.14} \\
    \midrule
    \rowcolor{black!7}
    \multicolumn{8}{l}{\textbf{TWIST2}} \\
    Original
      & \metricstat{67.42}{9.07} & \metricstat{47.21}{8.62} & \metricstat{37.48}{4.84} & \metricstat{249.49}{16.90}
      & \metricstat{120.62}{0.94} & \metricstat{14.06}{0.38} & \metricstat{0.00}{0.00} \\
    Ours (w/o H)
      & \metricstat{71.81}{9.35} & \metricstat{50.37}{8.94} & \metricstat{38.56}{5.02} & \metricstat{258.16}{17.42}
      & \metricstat{126.84}{1.28} & \metricstat{14.29}{0.41} & / \\
    Ours (w/ H)
      & \metricstat{79.64}{11.28}
      & \metricstat{102.58}{22.46}
      & \metricstat{40.31}{6.17}
      & \metricstat{275.83}{20.15}
      & \metricstat{143.75}{4.36}
      & \metricstat{14.68}{0.47}
      & \metricstat{2.18}{0.38} \\
    \midrule
    \rowcolor{black!7}
    \multicolumn{8}{l}{\textbf{BFM-Zero}} \\
    Original
      & \metricstat{79.13}{0.57} & \metricstat{67.55}{0.86} & \metricstat{41.49}{0.49} & \metricstat{293.60}{0.24}
      & \metricstat{156.37}{0.07} & \metricstat{8.02}{0.01} & \metricstat{0.00}{0.00} \\
    Ours (w/o H)
      & \metricstat{83.26}{0.74} & \metricstat{71.34}{1.12} & \metricstat{44.75}{0.63} & \metricstat{306.42}{1.36}
      & \metricstat{164.58}{0.93} & \metricstat{8.17}{0.03} & / \\
    Ours (w/ H)
      & \metricstat{95.48}{1.89}
      & \metricstat{139.72}{18.64}
      & \metricstat{51.08}{2.31}
      & \metricstat{326.81}{4.72}
      & \metricstat{186.93}{5.26}
      & \metricstat{8.46}{0.09}
      & \metricstat{2.16}{0.86} \\
    \bottomrule
  \end{tabular*}
\end{table*}

We evaluate our method in simulation and on real hardware, focusing on four questions:
\begin{itemize}
  \item \textbf{Q1:} Can our method improve human avoidance across different motion-tracking backbones?
  \item \textbf{Q2:} Can avoidance skills learned in simulation transfer to a real robot?
  \item \textbf{Q3:} Can our method outperform the baselines in both human avoidance and motion tracking?
  \item \textbf{Q4:} How do the key design choices affect avoidance and motion preservation?
\end{itemize}

\paragraph{Experimental Setup.}
\label{sec:experiment_setup}
We fine-tune SONIC~\cite{luo2026sonic}, TWIST2~\cite{ze2026twist2},
and BFM-Zero~\cite{li2026bfm} for human avoidance, starting from their officially released pretrained weights.
Fine-tuning is performed in Isaac Sim~\cite{NVIDIA_Isaac_Sim} and Isaac Gym~\cite{makoviychuk2021isaac} on a single NVIDIA GeForce RTX 4090 GPU in each case.
We select $500$ representative motion sequences that intrude into the surrounding space
from the BONES-SEED dataset for fine-tuning.
During training, humans are represented by fixed cylindrical and prismatic templates.
The LoRA adaptation rank is set to $16$.
The potential-field input $O_U$ uses $21$ body sampling points: $1$ on the pelvis and $10$ each on the arms and legs.
For real-world experiments, we deploy the fine-tuned SONIC policy on a Unitree G1 humanoid robot
and evaluate avoidance and motion tracking as nearby humans move and change poses.

\paragraph{Baselines.}
We compare three baselines.

\noindent\textbf{SONIC}~\cite{luo2026sonic} tracks reference motions without human perception or avoidance.

\noindent\textbf{CSF}~\cite{yang2026csf} rejects unsafe actions when risk is detected
and replaces the remaining reference with safe motions.

\noindent\textbf{ConstrainedMimic}~\cite{morton2026constrained}
uses kinematic and dynamic safety filters based on control barrier functions
to modify the reference motion and low-level control commands, respectively.

\paragraph{Metrics.}
\label{sec:evaluation_metrics}
SR is the percentage of test sequences completed without collisions or falls.
Tracking errors use the original references: $E_{\mathrm{g\text{-}mpbpe}}$ and $E_{\mathrm{mpbpe}}$
are global and root-relative mean body position errors in mm;
$E_{\mathrm{mpboe}}$ and $E_{\mathrm{mpjpe}}$ are mean body orientation and joint angle errors in $10^{-3}$ rad;
$E_{\mathrm{mpjve}}$ measures joint angle increment errors in $10^{-3}$ rad/frame~\cite{li2026haic}.
$E_{\mathrm{mpbpe}}^{a}$ and $E_{\mathrm{mpbpe}}^{o}$ evaluate limbs that penetrate human envelopes
under the original policy and the remaining body parts.
For scenes with penetration under the original policy, the normalized safety gain $\mathcal{M}$ is the increase in minimum signed distance over the sequence divided by the original penetration depth.

\subsection{Adaptation across Motion-Tracking Backbones}
\label{sec:cross_backbone}

To answer \textbf{Q1},
we apply the proposed avoidance pipeline to three pretrained backbones: SONIC, TWIST2, and BFM-Zero.
In Table~\ref{tab:cross_backbone}, Original denotes the pretrained policy,
while Ours w/o H and Ours w/ H denote the adapted policy evaluated without nearby humans
and in human proximity (H) scenes, respectively.

As shown in Table~\ref{tab:cross_backbone}, without nearby humans,
the tracking errors of Ours w/o H remain close to those of Original,
indicating preserved motion-tracking capabilities.
With humans nearby, $E_{\mathrm{mpbpe}}^{a}$ increases across all three backbones
as limbs at risk deviate to avoid collisions.
Smaller increases in $E_{\mathrm{mpbpe}}^{o}$ indicate that adjustments concentrate on these limbs,
while other body parts largely preserve their original motion trends.
The mean normalized safety gain $\mathcal{M}$ exceeds $1$ for all three backbones:
improvements in minimum signed distance exceed the original penetration depths on average after normalization.

\subsection{Real-World Deployment and Adaptation}
\label{sec:real_world}

\begin{figure*}[t]
  \centering
  \includegraphics[width=\textwidth]{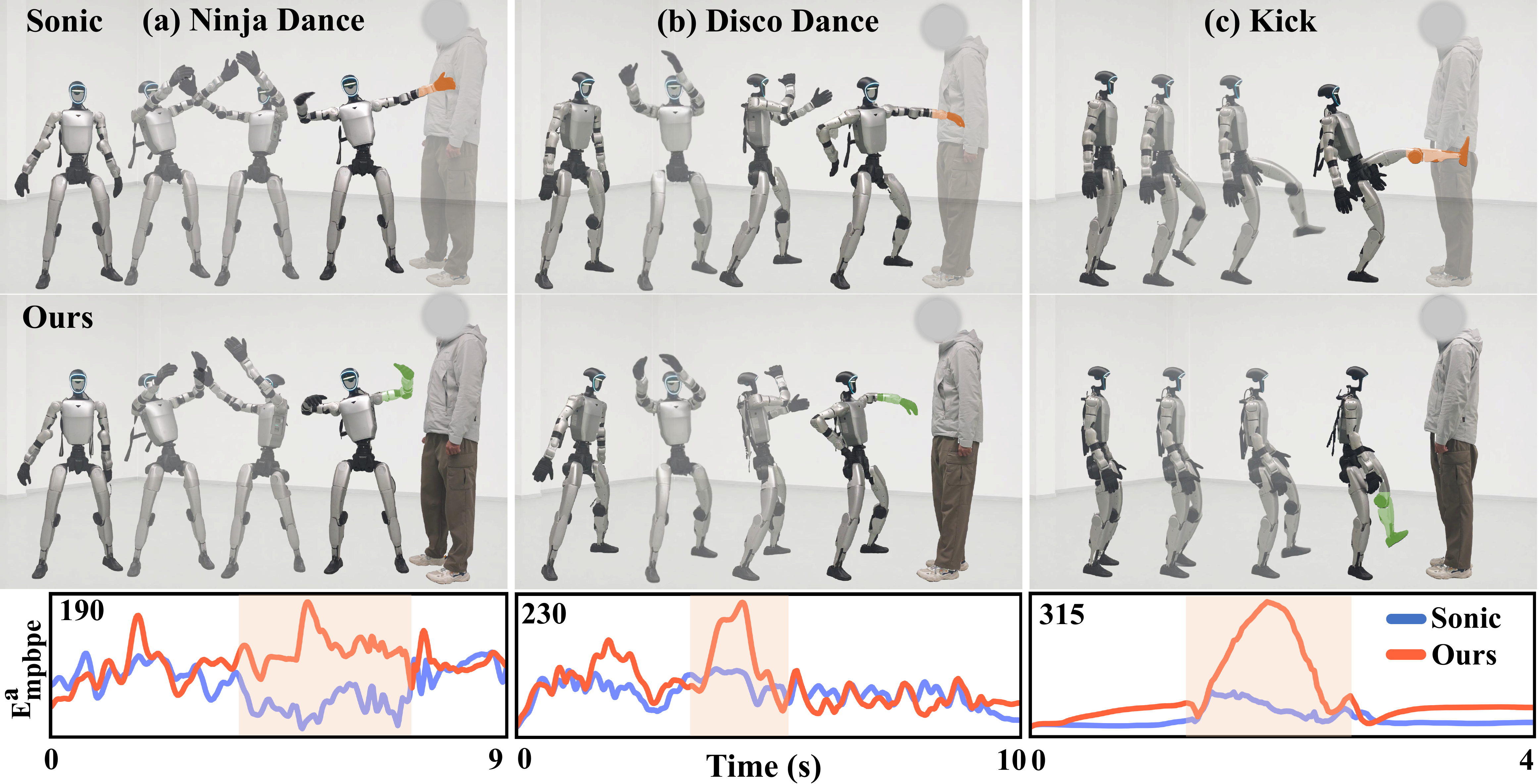}
  \caption{Real-world avoidance results for Ninja Dance, Disco Dance, and Kick.
  The top and bottom rows show the original SONIC policy and our policy, respectively.
  The bottom plots show the time series of $E_{\mathrm{mpbpe}}^{a}$ for the limbs involved in avoidance, with orange boxes marking the avoidance intervals.}
  \label{fig:real_world}
\end{figure*}

\begin{table}[t]
  \centering
  \caption{Real-world performance across four motion categories.}
  \label{tab:real_world}
  \footnotesize
  \setlength{\tabcolsep}{2pt}
  \renewcommand{\arraystretch}{1.10}
  \newcommand{\realstat}[2]{\ensuremath{#1_{\pm #2}}}
  \begin{tabular*}{\columnwidth}{@{\extracolsep{\fill}}lccc@{}}
    \toprule
    Motion
      & $E_{\mathrm{mpbpe}}^{a}$
      & $E_{\mathrm{mpbpe}}^{o}$
      & $\mathcal{M}$ \\
    \midrule
    Ninja Dance       & \realstat{148.97}{32.34} & \realstat{44.67}{14.09} & \realstat{2.78}{0.78} \\
    Disco Dance       & \realstat{70.65}{16.16} & \realstat{28.70}{6.59} & \realstat{2.10}{1.07} \\
    Kick              & \realstat{71.10}{37.05} & \realstat{38.67}{4.54} & \realstat{2.70}{1.25} \\
    Teleop            & \realstat{65.26}{11.21} & \realstat{36.63}{3.85} & \realstat{1.45}{0.63} \\
    \bottomrule
  \end{tabular*}
\end{table}

\begin{table}[t]
  \centering
  \caption{Comparisons and ablations on SONIC backbone.}
  \label{tab:comparison_ablation}
  \footnotesize
  \setlength{\tabcolsep}{2pt}
  \renewcommand{\arraystretch}{1.10}
  \newcommand{\comparisonstat}[2]{\ensuremath{#1_{\textcolor{black!50}{\pm #2}}}}
  \begin{tabular*}{\columnwidth}{@{\extracolsep{\fill}}lcccc@{}}
    \toprule
    Method
      & SR $\uparrow$
      & $E_{\mathrm{mpbpe}}^{a}\downarrow$
      & $E_{\mathrm{mpbpe}}^{o}\downarrow$
      & $\mathcal{M}$ \\
    \midrule
    SONIC
      & $0$
      & \comparisonstat{37.29}{0.42}
      & \comparisonstat{35.87}{1.79}
      & \comparisonstat{0.00}{0.00} \\
    CM
      & $83$
      & \comparisonstat{142.08}{20.03}
      & \comparisonstat{51.76}{12.07}
      & \cellcolor[HTML]{F3E6EF}\comparisonstat{1.27}{0.38} \\
    CSF
      & \cellcolor[HTML]{F3E6EF}$95$
      & \comparisonstat{172.18}{28.89}
      & \cellcolor[HTML]{F3E6EF}\comparisonstat{38.26}{10.21}
      & \comparisonstat{2.19}{0.45} \\
    \midrule
    w/o LoRA
      & $51$
      & \comparisonstat{110.84}{15.83}
      & \comparisonstat{68.64}{8.39}
      & \comparisonstat{2.04}{2.56} \\
    w/o PF
      & $60$
      & \comparisonstat{112.30}{14.64}
      & \comparisonstat{44.39}{15.01}
      & \comparisonstat{1.83}{2.17} \\
    w/o Flow
      & $79$
      & \cellcolor[HTML]{F3E6EF}\comparisonstat{89.66}{20.55}
      & \comparisonstat{50.66}{16.36}
      & \comparisonstat{1.99}{1.48} \\
    \midrule
    \textbf{Ours}
      & \cellcolor[HTML]{E4C7DC}$\mathbf{100}$
      & \cellcolor[HTML]{E4C7DC}\comparisonstat{\mathbf{84.03}}{19.30}
      & \cellcolor[HTML]{E4C7DC}\comparisonstat{\mathbf{37.75}}{7.51}
      & \cellcolor[HTML]{E4C7DC}\comparisonstat{\mathbf{1.21}}{0.14} \\
    \bottomrule
  \end{tabular*}
\end{table}

To answer \textbf{Q2}, we deploy SONIC fine-tuned in simulation on Unitree G1.
We evaluate Ninja Dance, Disco Dance, Kick, and Teleop
for avoidance during dynamic motions, whole-body coordination, and teleoperation.

Under approximately $30\%$ body occlusion and human speeds up to $2$ m/s,
mean human-center localization error is $8$ cm.
Evaluated against SMPL~\cite{loper2023smpl} surfaces reconstructed by EasyMocap~\cite{shuai2022novel},
our envelopes achieve $93\%$ mean surface coverage, with uncovered regions extending at most $2$ cm beyond the boundary.
Avoidance commands follow detected human motion changes with an estimated mean delay of $100$ ms.
Our policy achieves $100\%$ SR over $20$ real-world trials for each of the four motions,
with a minimum clearance of $3$ cm from the SMPL surfaces.
Tracking errors increase on hardware (Tables~\ref{tab:cross_backbone} and~\ref{tab:real_world}),
but local avoidance and overall motion trends are preserved.

The top row of Fig.~\ref{fig:real_world} shows the original policy,
where limbs approach nearby humans and risk collisions in all three motions.
With the same references, our policy in the bottom row avoids humans through local adjustments
while preserving the original motion trends.
The same policy handles arm avoidance during dance and leg avoidance during kicking.
In the bottom plots, tracking errors are close to those of the original policy before avoidance,
briefly increase during avoidance (orange boxes), and rapidly decline as reference tracking resumes.

\begin{figure}[!t]
  \centering
  \includegraphics[width=0.92\columnwidth]{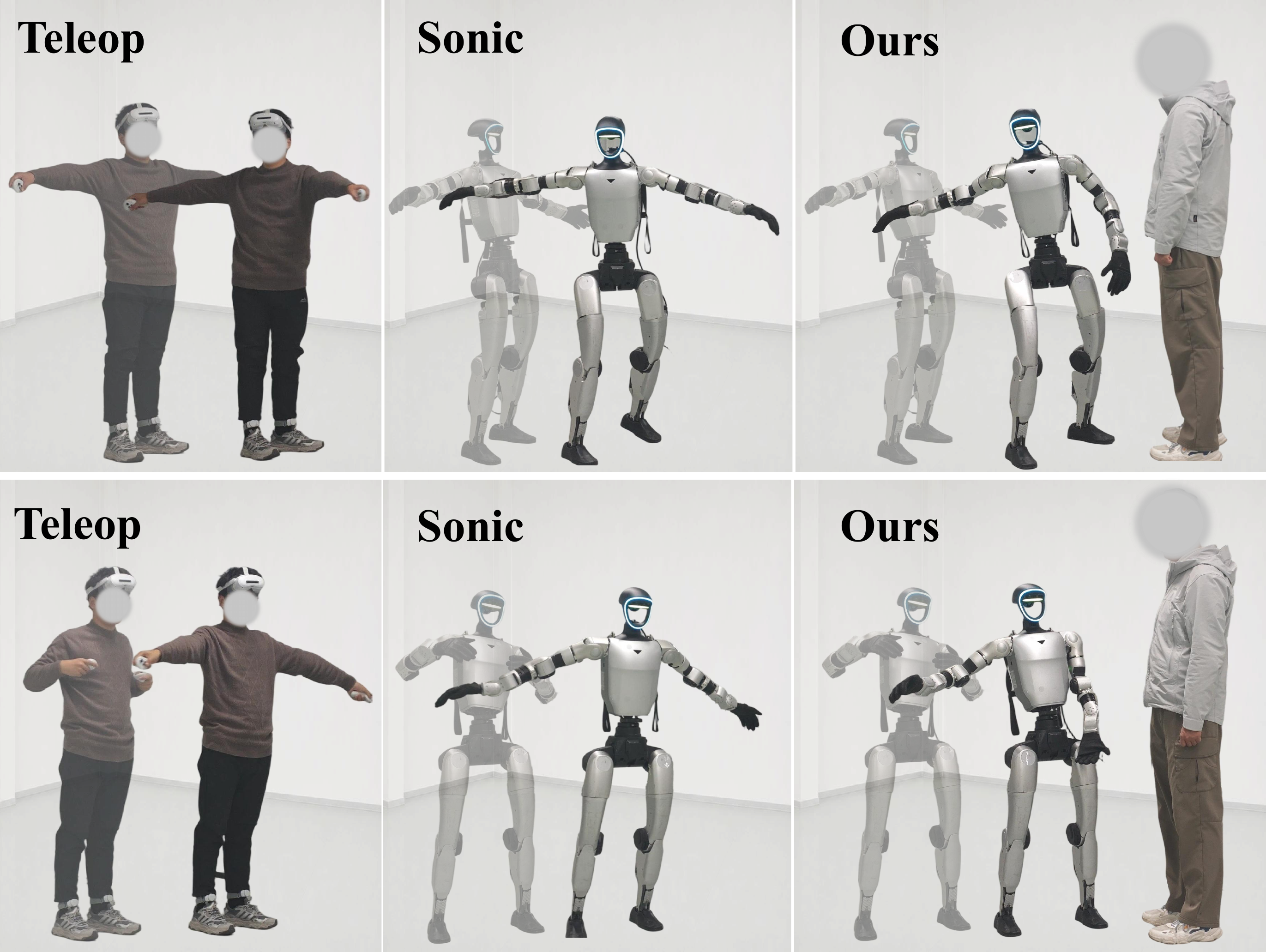}
  \caption{Local avoidance and motion preservation during teleoperation.
  Columns show the operator, the original SONIC policy, and our policy, respectively.}
  \label{fig:teleop}
  \par\smallskip
  \includegraphics[width=\columnwidth]{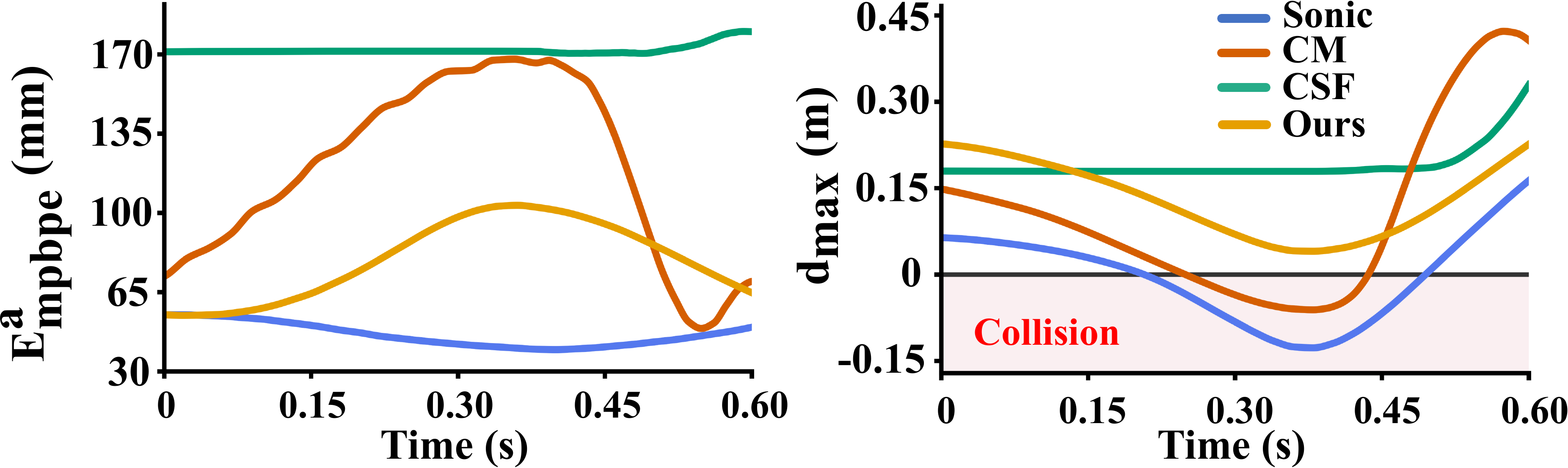}
  \caption{Tracking error and human clearance during arm swings.}
  \label{fig:baseline2}
\end{figure}

\begin{figure}[t]
  \centering
  \includegraphics[width=\columnwidth]{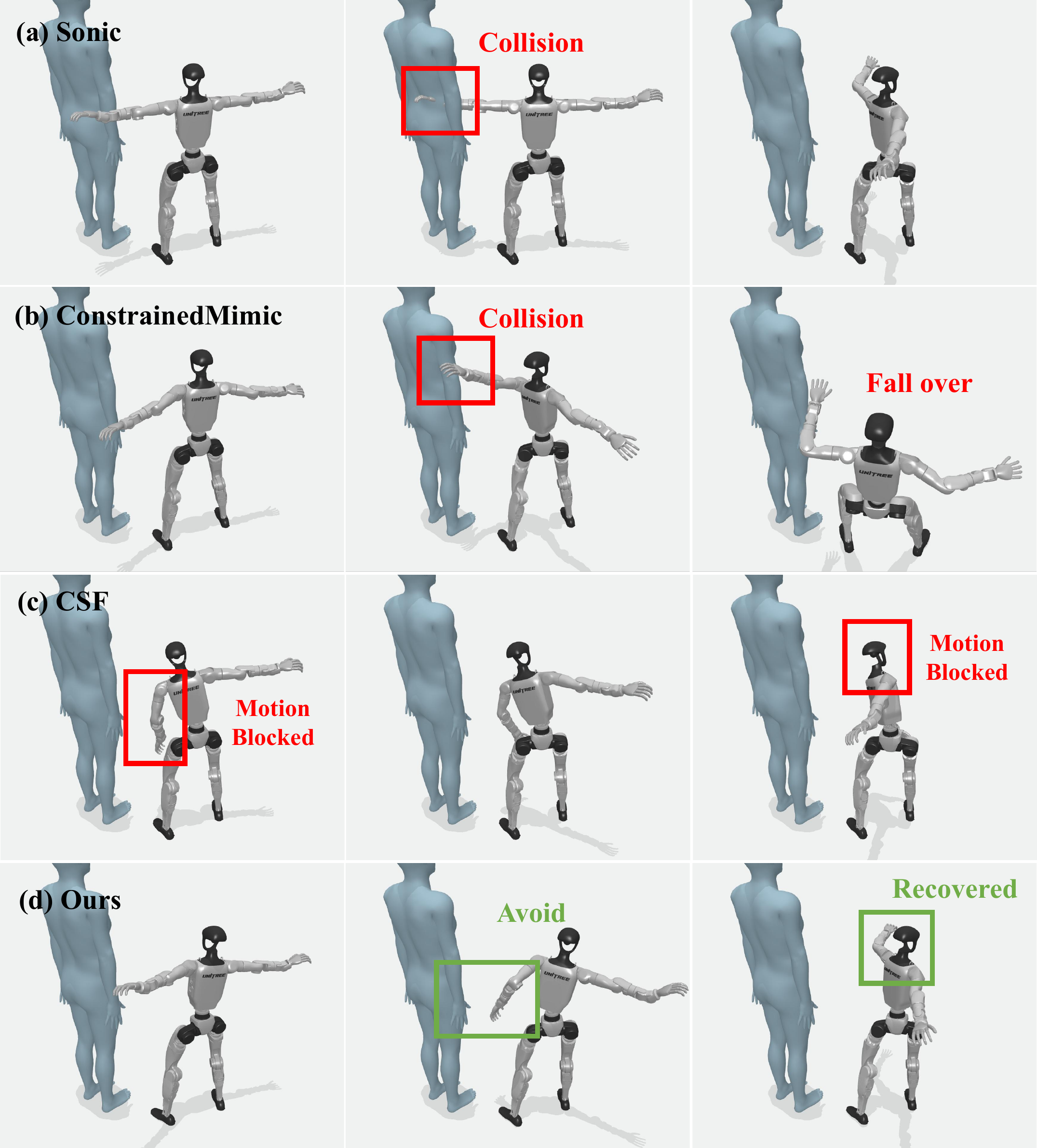}
  \caption{Comparison of avoidance and motion preservation across four methods in an arm-swing task.}
  \label{fig:arm_swing_montage}
\end{figure}

Figure~\ref{fig:teleop} shows diverse teleoperated motions.
As the operator moves in real time, our policy adjusts limbs near humans to avoid them,
while unaffected limbs continue to track the operator, preserving overall motion coordination.

Table~\ref{tab:real_world} reports the means and sample standard deviations of the execution metrics over 20 trials per task.
Across all four tasks, $E_{\mathrm{mpbpe}}^{a}$ exceeds $E_{\mathrm{mpbpe}}^{o}$,
while tracking errors remain low in the remaining body parts,
indicating localized avoidance deviations.
The normalized safety gain $\mathcal{M}$ reflects the retained safety margin.

\subsection{Motion Preservation during Avoidance}
\label{sec:motion_preservation}

To answer \textbf{Q3}, we compare our method with original SONIC, ConstrainedMimic (CM), and CSF on the SONIC backbone.
A human enters the region swept by the reference motion.
All four methods use the same reference and human trajectories and initial state
to evaluate motion preservation during avoidance.

Figure~\ref{fig:arm_swing_montage} illustrates motion preservation during avoidance.
CM corrects the current motion but leaves insufficient clearance for subsequent motion, leading to a collision and fall.
CSF blocks motions toward the human and subsequent motions, truncating the reference sequence.
Our method makes local adjustments that preserve the overall motion trends, then resumes reference tracking.

Figure~\ref{fig:baseline2} plots tracking error and human clearance throughout execution.
Our tracking error rises during avoidance and then falls, while human clearance remains positive.
SONIC and CM enter the collision region, while CSF maintains high tracking error.
Table~\ref{tab:comparison_ablation} summarizes overall performance over 100 trials per task:
our $E_{\mathrm{mpbpe}}^{a}$ and $E_{\mathrm{mpbpe}}^{o}$ values are lower than those of CM and CSF, indicating better motion preservation
in both affected and remaining body parts.
A higher success rate and $\mathcal{M}$ closer to the contact threshold of $1$
indicate more reliable collision-free execution with less excessive avoidance.

\subsection{Ablation Studies}
\label{sec:ablation}

To answer \textbf{Q4}, we examine the roles of conditional LoRA adaptation,
the potential-field input representation, and the velocity-flow tracking reward on the SONIC backbone,
using the three ablation settings in Table~\ref{tab:comparison_ablation}.
All variants use the same pretrained policy, fine-tuning data, interaction budget, and fixed test set.

\begin{figure}[t]
  \centering
  \includegraphics[width=\columnwidth]{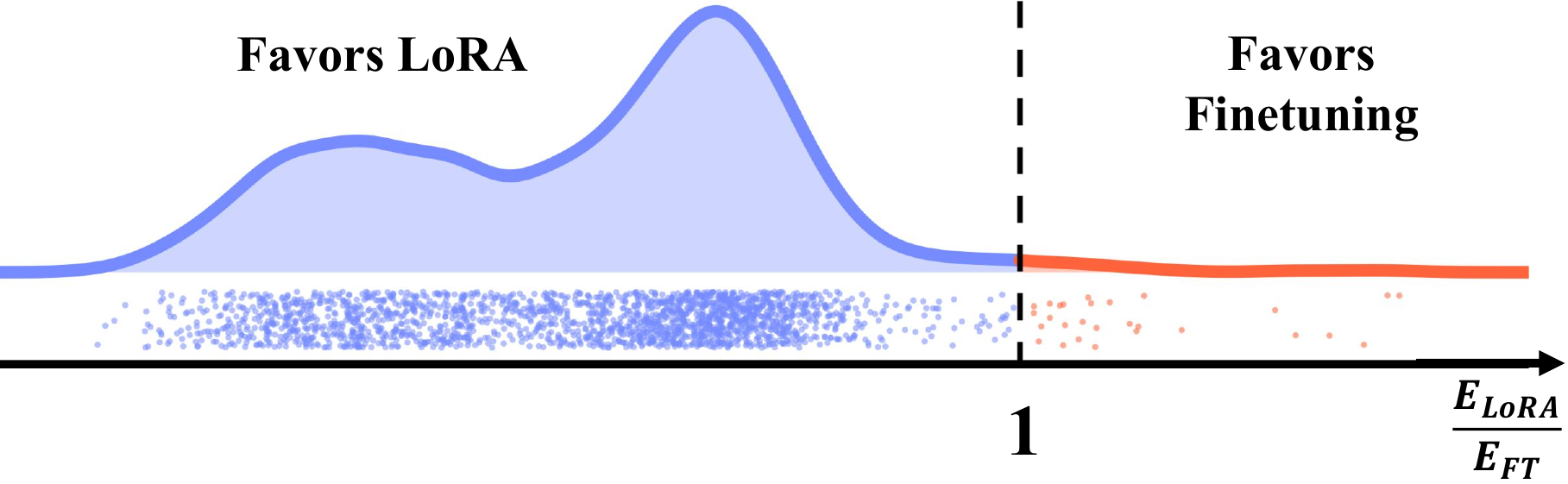}
  \caption{Tracking-error ratios on unseen motions.
  Each ratio compares $E_{\mathrm{mpbpe}}$ under conditional LoRA to full-parameter fine-tuning;
  the dashed line marks $1$.}
  \label{fig:ablation_lora}
\end{figure}

\paragraph{Conditional LoRA.}
We replace conditional LoRA with full-parameter fine-tuning (w/o LoRA in Table~\ref{tab:comparison_ablation}).
Conditional LoRA achieves more reliable avoidance while keeping both affected and remaining body parts closer to the reference motion.
This indicates that low-rank adaptation limits disruption to the original motions while learning local corrections for human avoidance.

We also evaluate tracking without nearby humans on $2048$ motions excluded from fine-tuning.
Figure~\ref{fig:ablation_lora} reports the $E_{\mathrm{mpbpe}}$ ratio of conditional LoRA to full-parameter fine-tuning.
Most ratios are below $1$, indicating that conditional LoRA better preserves tracking accuracy and generalization on unseen motions.

\begin{figure}[t]
  \centering
  \includegraphics[width=\columnwidth]{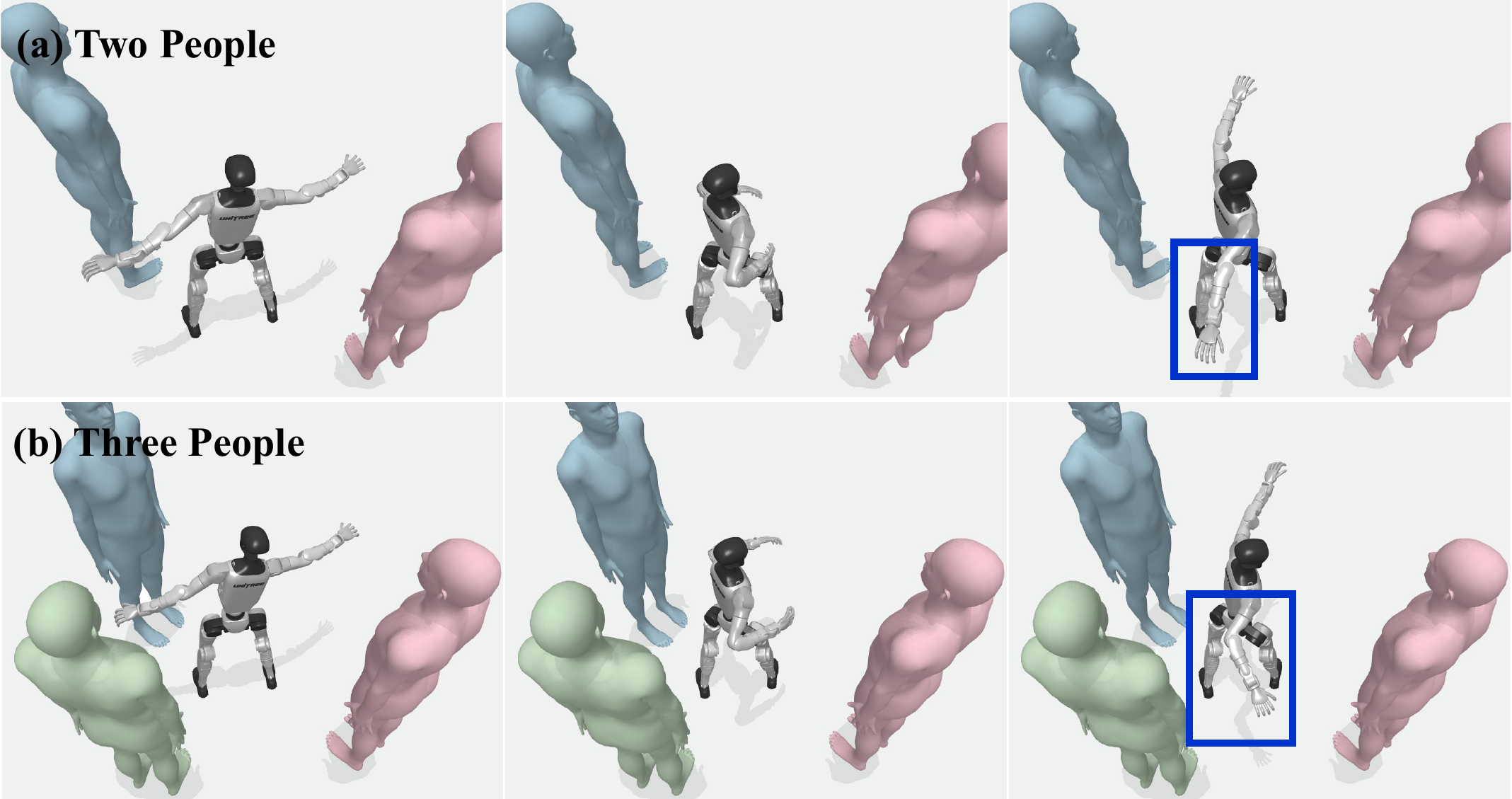}
  \caption{Local avoidance with two (top) and three (bottom) humans.
  Columns show three successive instants.
  Blue boxes highlight local arm adjustments.}
  \label{fig:multi_human_potential}
\end{figure}

\paragraph{Potential-Field Representation.}
We replace $O_U$ with distances and directions to human centers (w/o PF in Table~\ref{tab:comparison_ablation}).
Potential-field inputs improve avoidance and tracking, with $\mathcal{M}$ nearer $1$,
indicating better coordination of avoidance magnitude and direction with fewer deviations.

Figure~\ref{fig:multi_human_potential} shows local avoidance among multiple humans while preserving overall motion.
Each sample encodes combined potential energy, steepest-descent direction,
and normal and tangential flow components.
By superposing human potentials, we represent varying human counts and arrangements with fixed-size inputs,
enabling multi-human generalization without changing the network size.

\paragraph{Flow Reward.}
Removing the velocity-flow tracking reward (w/o Flow in Table~\ref{tab:comparison_ablation})
lowers success and increases tracking errors in both body groups.
The higher $\mathcal{M}$ indicates a larger normalized avoidance margin.
The target flow guides avoidance direction and magnitude while preserving reference-motion components tangent to equipotential surfaces.
This guidance coordinates avoidance with tracking while limiting excessive corrections and interference with other limbs.
Without the directional guidance provided by this reward, training also converges more slowly.

\section{Conclusion}
\label{sec:conclusion}

We introduce ALERT, a pipeline that extends pretrained humanoid motion trackers with human avoidance.
LiDAR-based human envelopes define repulsive potentials, whose velocity flows guide conditional low-rank adaptation.
Experiments across three backbones and on Unitree G1 demonstrate local avoidance, rapid tracking recovery,
and preserved tracking of unseen motions.
ALERT outperforms CM and CSF in success rate and tracking accuracy.

However, convex envelopes cannot represent gaps between human limbs,
potentially making close-range avoidance conservative.
Future work will incorporate articulated human geometry for finer local avoidance.


{
    \small
    \bibliographystyle{ieeenat_fullname_citeorder}
    \bibliography{main}
}


\end{document}